\documentclass[english]{article}
\usepackage[T1]{fontenc}
\usepackage[latin9]{inputenc}
\usepackage{geometry}
\usepackage{amstext}
\usepackage{graphicx}
\usepackage{setspace}

\usepackage{babel}
\begin{document}

\title{A statistical model for aggregating judgments by incorporating peer
predictions}

\author{John McCoy$^{3}$ and Drazen Prelec$^{1,2,3}$ \\
$^{1}$Sloan School of Management\\
 Departments of $^{2}$Economics, and $^{3}$Brain \& Cognitive Sciences\\
Massachusetts Institute of Technology, Cambridge MA 02139\\
jmccoy@mit.edu, dprelec@mit.edu}
\maketitle
\begin{abstract}
We propose a probabilistic model to aggregate the answers of respondents
answering multiple-choice questions. The model does not assume that
everyone has access to the same information, and so does not assume
that the consensus answer is correct. Instead, it infers the most
probable world state, even if only a minority vote for it. Each respondent
is modeled as receiving a signal contingent on the actual world state,
and as using this signal to both determine their own answer and predict
the answers given by others. By incorporating respondent's predictions
of others' answers, the model infers latent parameters corresponding
to the prior over world states and the probability of different signals
being received in all possible world states, including counterfactual
ones. Unlike other probabilistic models for aggregation, our model
applies to both single and multiple questions, in which case it estimates
each respondent's expertise. The model shows good performance, compared
to a number of other probabilistic models, on data from seven studies
covering different types of expertise. 
\end{abstract}

\section*{Introduction}

It is a truism that the knowledge of groups of people, particularly
experts, outperforms that of individuals \cite{surowiecki2005wisdom}
and there is increasing call to use the dispersed judgments of the
crowd in policy making \cite{sunstein2006infotopia}. There is a large
literature spanning multiple disciplines on methods for aggregating
beliefs (for reviews see \cite{cooke1991experts,clemen1999combining,clemen2007aggregating}),
and previous applications have included political and economic forecasting
\cite{budescu2014identifying,mellers2014psychological}, evaluating
nuclear safety \cite{cooke2008tu} and public policy \cite{morgan2014use},
and assessing the quality of chemical probes \cite{oprea2009crowdsourcing}.
However, previous approaches to aggregating beliefs have implicitly
assumed `kind' (as opposed to `wicked') environments \cite{hertwig2012tapping}.
In a previous paper, \cite{psm} we proposed an algorithm for aggregating
beliefs using not only respondent's answers but also their prediction
of the answer distribution, and proved that for an infinite number
of non-noisy Bayesian respondents, it would always determine the correct
answer if sufficient evidence was available in the world. 

Here, we build on this approach but treat the aggregation problem
as one of statistical inference. We propose a model of how people
formulate their own judgments and predict the distribution of the
judgments of others, and use this model to infer the most probable
world state giving rise to the observed data from people. The model
can be applied at the level of a single question but also across multiple
questions, to infer the domain expertise of respondents. The model
is thus broader in scope than other machine learning models for aggregation
in that it accepts unique questions, but can also be compared to their
performance across multiple questions. We do not assume that the aggregation
model has access to correct answers or to historical data about the
performance of respondents on similar questions. By using a simple
model of how people make such judgments, we are able to increase the
accuracy of the group's aggregate answer in domains ranging from estimating
art prices to diagnosing skin lesions. 

\subsection*{Possible worlds and peer predictions}

Condorcet's Jury Theorem \cite{condorcet1785essay}, of 1785, considers
a group of people making a binary choice with one option better for
all members of the group. All individuals in the group are assumed
to vote for the better option with probability $p>0.5$. This is also
known as the assumption of voter competence. The theorem states that
the probability that the majority vote for the better alternative
exceeds $p$, and approaches $1$ as the group size increases to infinity.\footnote{See \cite{sunstein2006infotopia} for a review of modern extensions
to Condorcet's Jury Theorem.}. Following this theorem, much of the belief aggregation and social
choice theory literature assumes voter competence and thus focusses
on methods which use the group's consensus answer, for example by
computing the modal answer for categorical questions or the mean or
median answer for aggregating continuous quantities. \footnote{The work on aggregating continuous quantities using the mean or median
also has an interesting history, dating back to Galton estimating
the weight of an ox using the crowd's judgments \cite{galton1907vox}. }

To build intution both how our model relates to, but also departs
from, previous work on aggregating individual judgments, we begin
with a model of Austen-Smith and Banks \cite{austen1996information}
who argue that previous work following the Concordet Jury Theorem
starts the analysis `in the middle' with members of the group voting
according to the posterior probabilities that they assign to the various
options, but without the inputs to these posteriors specified. They
describe a model with two possible world states $\Omega\in\{A,B\}$
and two possible options $\{A,B\}$ to select and assume that everyone
attaches a utility of 1 to the option that is the same as the actual
world state, and a utility of 0 to the other option. Throughout, we
conceptualise the possible world states as corresponding to the different
possible answers to the question under consideration, with the correct
answer, based on current evidence, corresponding to the actual or
true state of the world.\footnote{In this paper, we ignore utilities and assume that people vote for
the world state that they believe is most probable. The inputs we
elicit from respondents are suffficient to implement mechanisms that
make answering in this way incentive compatible, for example by using
the Bayesian Truth Serum \cite{prelec2004bayesian}\cite{john2012measuring}.} The actual world state is unknown to all individuals, but there is
a common prior $\pi\in[0,1]$ that the world is in state A, and each
individual receives a private signal $s\in\left\{ a,b\right\} $.\footnote{Throughout the paper, we describe other models using a consistent
notation, rather than necessarily that used by the original authors. } Crucially, Austen-Smith and Banks assume that signal $a$ is strictly
more likely than signal $b$ in world $A$ and signal $b$ is strictly
more likely than signal $a$ in world $B$. That is, they assume that
the distribution on signals is such that signal $a$ has probability
greater than $0.5$ in world $A$ and signal $b$ has probability
greater than 0.5 in world $B$. After receiving her signal, each individual
updates her prior belief and votes for the alternative she believes
provides the higher utility. 

In the Austen-Smith and Banks model, the majority verdict favors the
most widely available signal, but the assumption that the most widely
available signal is correct may be false for actual questions. For
example, shallow information is often widely available but more specialized
information which points in a different (correct) direction may be
available only to a few experts. The model thus does not apply to
situations where the majority is incorrect. 

We previously proposed a model that does not assume that the most
widely available information is always correct \cite{psm}. Using
the same setup as Austen-Smith and Banks, we require only that signal
$i$ is more probable in world $I$ than in the other world, This
i does not imply that signal $i$ is necessarily the most likely signal
in world $I$. If the world is in state $I$ then we refer to signal
$i$ as the correct signal since a Bayesian respondent receiving signal
$i$ places more probability on the correct world $i$ than a respondent
receiving any other signal. Our model does not assume that in any
given world state the most likely signal is also the correct signal.
For example, under our model, signal $a$ may have probability 0.8
in state $A$ and probability 0.7 in state $B$ and thus be more likely
than signal $b$ in both world states. Under this signal distribution,
if the actual state was $B$ then the majority would receive signal
$a$, and, assuming a uniform common prior, would vote for state $A$
and so would be incorrect.\footnote{Assuming this signal distribution and a uniform common prior, consider
a respondent who received signal $s=a$. Then, their estimate of the
probability that the world is in state $A$ is $p(\Omega=A|s=a)=p(s=a|\Omega=A)p(\Omega=A)/p(s=a)=(.8)(.5)/((.8)(.5)+(.7)(.5))$
and since this quantity is higher than 0.5, respondents receiving
signal $a$ vote that the world is most likely in state $A$. But
if the actual world is state $B$, respondents have a .7 probability
of receiving signal $a$ and hence the majority of respondents will
vote incorrectly. } 

We call the model proposed in this paper the `possible worlds model'
(PWM) since to determine the actual world state it is not sufficient
to consider only how people actually vote, but, instead, one needs
to also consider the distribution of votes in all possible worlds.
That is, to determine the correct answer, one requires not simply
what fraction of individuals voted for a particular answer under the
actual world state, but also what fraction would have voted for that
answer in all counterfactual world states. A useful intuition is that
an answer which is likely in all possible worlds should be discounted
relative to an answer which is likely only in one world state: simply
because 60\% of respondents vote for option $a$ does not guarantee
that the world is actually in state $A$ since perhaps 80\% of respondents
would have voted for $a$ if this was actually the case. If an aggregation
algorithm had access to the world prior and signal distributions for
every possible world, then recovering the correct answer would be
as simple as determining which world had an expected distribution
of votes matching the observed frequency of votes.\footnote{The algorithm would also require access to the world prior to determine
the expected vote frequencies, which may not match the expected signal
frequencies. } The new challenge presented by our possible worlds model is how to
tap respondent's meta-knowledge of such counterfactual worlds, and
our solution is to do this by eliciting respondent's predictions of
how other respondents vote. We model such predictions as depending
on both a respondent's received signal and their knowledge of the
signal distribution and world prior, which allows us to infer the
necesary information about counterfactual worlds.\footnote{In a previous paper \cite{psm} we showed that given our weaker assumptions
on the signal distribution, eliciting only votes or even full posteriors
over worlds from ideal Bayesian respondents is not sufficient to identify
the actual world for multiple choice questions with any number of
options. On the other hand, we provide formal conditions under which
eliciting votes and predictions about others votes will always uncover
the correct answer. For binary questions, the answer which is more
popular than predicted can be shown to be the best answer under such
conditions.} The probabilistic generative model proposed in this paper assumes
that such predictions are generated by respondents as noisy approximations
to an exact Bayesian computation. 

Our assumptions about people's predictions of others are supported
by work in psychology that has robustly demonstrated that people's
predictions of others answers relate to their own answer, and, in
particular, are consistent with respondents implicitly conditioning
on their own answer as an `informative sample of one'' \cite{dawes1989statistical,dawes1990false,hoch1987perceived,krueger1994truly,robbins2005social}.
People show a so-called false consensus effect whereby people who
endorse an answer believe that others are also more likely to endorse
it. For example, in one study \cite{ross1977false}, about 50\% of
surveyed undergraduates said that they themselves would wear a sign
saying ``Repent'' around campus for a psychology experiment and
predicted that 61\% of other students would wear such a sign, whereas
students who would not wear such a sign predicted that 30\% would.
As Dawes pointed out in a seminal paper \cite{dawes1989statistical}
there is nothing necessarily false about such an effect, it is rational
to use your own belief as a sample of the population and base your
prediction of others on your own beliefs. In our model, people's beliefs
do not directly affect their predictions, but rather both their own
beliefs and predictions are both conditional on the private signal
which they received. In this paper, we use the model discussed above
to develop statistical models of aggregation. We thus turn now to
previous attempts that tackle the aggregation problem as one of Bayesian
inference, attempting to infer the correct answer given data from
multiple individuals.

\subsection*{Aggregation as Bayesian inference}

One approach to the problem of aggregating the knowledge or answers
of a group of people is to treat it as a Bayesian inference problem,
and model how the data elicited from respondents is generated conditional
on the actual world state. The observed data is then used to infer
a posterior distribution over possible world states. Unlike our model,
other such aggregation models either require data from respondents
answering multiple questions or require the user to specify a prior
distribution over the answers.

Bayesian approaches to aggregating opinions are not new \cite{winkler1968consensus,morris1977combining}.
Using such an approach, the aggregator specifies a prior distribution
over the variable of interest and updates this prior with respect
to a likelihood function associated with information from respondents
about this variable \cite{cooke1991experts}. For example, Bayesian
models with different priors and likelihood functions have been compared
for the problem of determining the value of an indicator variable
where each expert gives the probability that the indicator variable
is turned on \cite{clemen1990unanimity}. 

We will compare the PWM to two other hierarchical Bayesian models
for aggregation: a Bayesian Cultural Consensus Theory model \cite{karabatsos2003markov,oravecz2014bayesian}
and a Bayesian cognitive hierarchy model \cite{lee2014using}. We
give formal definitions of these two models in section \ref{sec:Comparison-models},
and explain here how they relate to other models for aggregating opinions. 

Cultural consensus models \cite{romney1986culture,batchelder1988test,romney1999cultural,weller2007cultural}
are a class of models used to uncover the shared beliefs of a group.
Respondents are asked true or false questions, and these answers are
used to infer each respondent's cultural competence and the culturally
correct consensus answers. The standard cultural consensus model,
called the General Concordet Model \cite{romney1986culture}, assumes
that, depending on question difficulty and respondent competence,
people either know the correct answer or make a guess. The Bayesian
cultural consensus model \cite{karabatsos2003markov,oravecz2014bayesian}
is a hierarchical Bayesian model that adds hyperpriors and a noise
model to the General Concordet Model.

Outside of cultural consensus theory, a number of hierarchical Bayesian
models for aggregation have been developed that attempt to model how
people produce their answers given some latent knowledge, and then
aggregate information at the level of this latent knowledge. For example,
when playing ``The Price is Right'' game show, people's bids for
a product may not correspond to their knowledge of how much the product
is worth (because of the competive nature of the show), and so infering
their latent knowledge and aggregating at this level will give more
accurate estimates about the product than aggregating at the level
of their bids \cite{lee2011wisdom}. Such hierarchical Bayesian models
also include models for aggregating over multidimension stimuli, for
example combinatorial problems \cite{yi2012wisdom} and travelling
salesman problems \cite{yi2010wisdom}. Modeling the cognitive processes
behind someone's answer also allows individual hetrogeneity to be
estimated, for example the differing knowledge that more or less expert
respondents have in ranking tasks \cite{lee2012inferring} or using
a cognitive hierarchy model to account for people's differing levels
of noise and calibration when respondent's are answering binary questions
using probabilties.\cite{lee2014using,turner2014forecast} . 

The models discussed above reflect important advances for solving
the problem of belief aggregation, but they remain focussed on attempting
to derive the consensus answer, allowing for differences in how this
consensus answer is represented and reported. \footnote{One of the clearest expressions of this is from \cite{lee2014using},
``A more concrete and perhaps more satisfying answer is that the
model works by identifying agreement between the participants in the
high-dimensional space defined by the 40 questions. Thinking of each
person\textquoteright s answer as a point in a 40-dimensional space
makes it clear that, if a number of people give similar answers and
so correspond to nearby points, it is unlikely to have happened by
chance. Instead, these people must be reflecting a common underlying
information source. It then follows that a good wisdom of the crowds
answer is near this collection of answers, people who are close to
the crowd answer are more expert, and people who need systematic distortion
of their answers to come close to this point must be miscalibrated.
Based on these intuitions, the key requirement for our model to perform
well is that a significant number of people give answers that contain
the signal of a common information source.''}

Previously\cite{psm}, we described a simple principle for aggregating
answers to binary questions based on the possible worlds model: select
the answer which is more popular than people predict, rather than
simply the most popular answer. While this principle has the advantage
of simplicity, developing a probabilistic generative model based on
the possible worlds model has several benefits. First, such a generative
model yields a posterior distribution over world states, rather than
simply selecting an answer. Second, generative models allow us to
easily incorporate other kinds of information into the model. For
example, they facilitate the introduction of latent expertise variables
allowing the model to be run across multiple questions. Third, the
generative model allows one to explicitly model the noise in respondent's
beliefs, rather than assuming that respondents are perfect Bayesians.
Fourth, as discussed later, the generative model lends itself to a
number of possible extensions.

\section*{A generative possible worlds model }

\subsection*{A model for single questions}

\begin{figure}[p]
\includegraphics[scale=0.8]{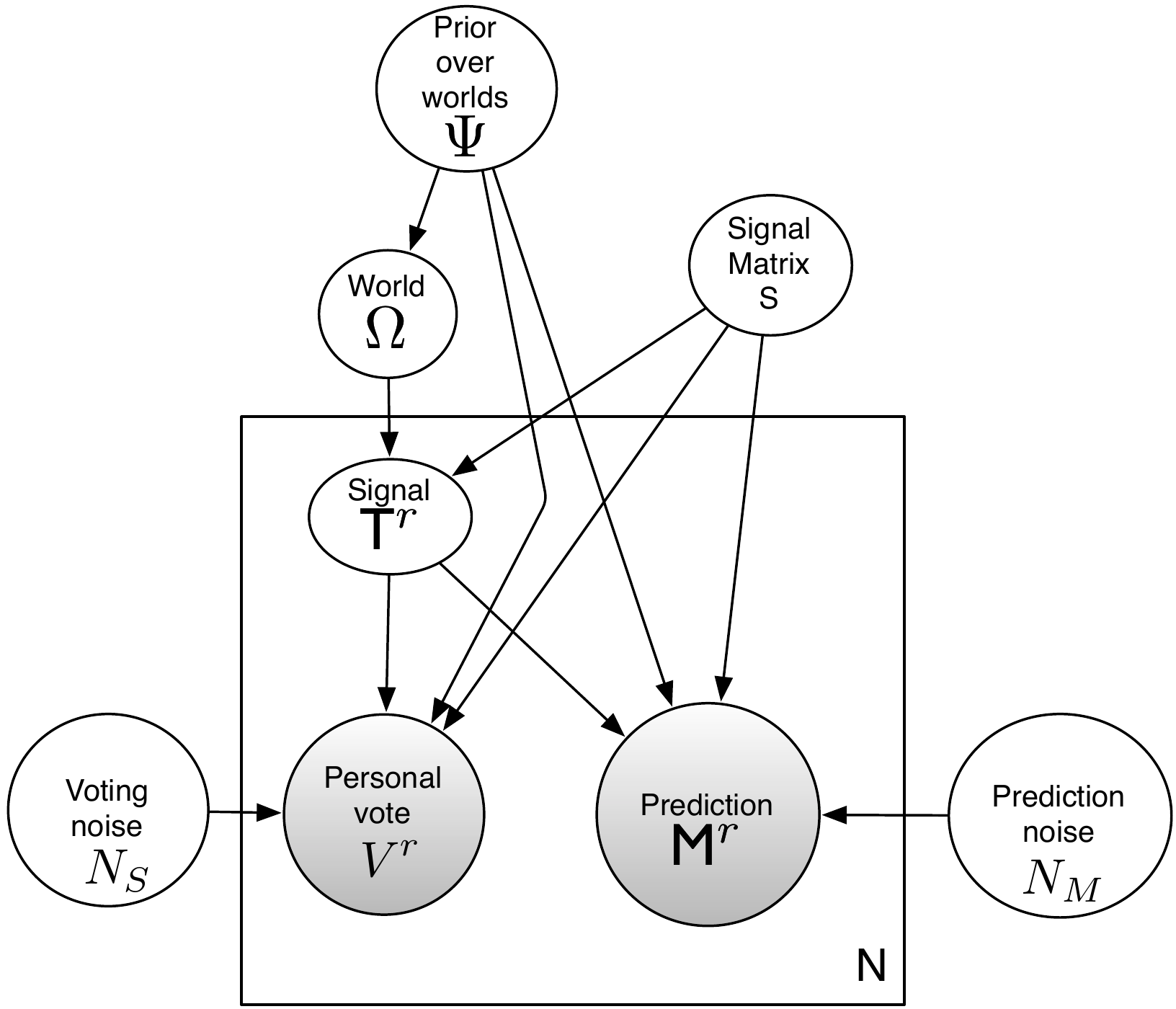}

\caption{The single question possible worlds model (PWM) which is used to infer
the underlying world state based on a group's votes and predictions
of the votes of others. In keeping with standard graphical model plate
notation \cite{jordan2004graphical,kollar2009probabilistic}, nodes
are random variables, shaded nodes are observed, an arrow from node
$X$ to node $Y$ denotes that $Y$ is conditionally dependent on
$X$, a rectangle around variables indicates that the variables are
repeated as many times as indicated in the lower right corner of the
rectangle.\label{fig:basic_graphical_model-1}}
\end{figure}

In this section we discuss applying the model to single questions
and in the next section we discuss inferring respondent level parameters
by applying the model across multiple questions. We present the core
model for binary questions here and return in the discussion section
to our modeling assumptions, and consider alternatives and extensions,
for example to non-binary multiple choice questions. The graphical
model \cite{jordan2004graphical,kollar2009probabilistic} for respondents
voting on a single questions is shown in Figure \ref{fig:basic_graphical_model-1}.
Suppose that $N$ respondents each give their personal answer to a
multiple choice question with two possible answers and predict the
distribution of votes given by other respondents. We assume that each
of the possible answers to the question corresponds to a different
underlying world state $\Omega\in\left\{ A,B\right\} $ and denote
the actual state of the world as $\Omega^{*}$. For example, if respondents
are asked, ``Is Philadelphia the capital of Pennsylvania?'' the
two possible worlds states are true and false with false being the
actual world state, since the capital of Pennyslvania is actually
Harrisburgh. A respondent $r$ indicates their vote $V^{r}$ for the
answer that they believe is most likely, with superscripts indexing
a particular respondent throughout. Respondents also give their prediction
$M^{r}$ (`$M$' for meta-prediction) of the fraction of respondent's
voting for each answer. For binary questions, $M^{r}$ is completely
specified by the fraction of people predicted to vote for option $A$.
Respondents dmetimes additionally asked to indicate the confidence
that they have in their answer being correct, and in section \ref{subsec:model_extensions}
we develop a model that uses such confidence judgments.

The prior over worlds represents information that is common knowledge
among all respondents. It is given by a binomial distribution with
parameter $\psi$ and the hyperprior over $\psi$ is a uniform Beta
distribution. Each respondent is assumed to receive a private signal
that represents additional information above the common knowledge
information captured by the prior, with respondent $r$ receiving
signal $T^{r}$. We assume that there are the same number of possible
signals as there are possible world states with $T^{r}\in\{{a,b}\}$
for binary questions. The signal that a respondent receives is determined
by the actual world $\Omega^{*}$ and a signal distribution. The signal
distribution $S$ is represented in the binary case as a $2\times2$
left stochastic matrix (i.e. the columns sum to 1) where the $iJ$-th
entry corresponds to the probability of an arbitrary respondent receiving
signal $i$ when $\Omega^{*}=J$ (assuming a fixed ordering of world
states and signals). In other words, each respondent's signal is sampled
from a categorical distribution with the $\Omega^{*}$-th column of
the signal distribution matrix giving the probabilities of the different
signals. Note that the model for single questions assumes that respondents
are identical except for the signal that they receive, and that for
any given signal all respondents have the same probability of receiving
it.

The prior we specify over the signal distribution does not impose
the substantive assumption that if $\Omega^{*}=i$ then signal $i$
is necessarily the most probable signal. Instead, we allow the possibility
that respondents receiving an incorrect signal may be the majority.
Specifically, we assume that the signal distribution is sampled uniformly
from the set of left stochastic matrices with the constraint that
$p(T^{r}=i|\Omega=i)>p(T^{r}=i|\Omega=j)$ for all $i,j\neq i$. This
prior provides a constraint on the world in which a given signal is
more likely, not a constraint on which signal is more likely for a
given world state, and does not imply that the majority of respondents
receive a signal corresponding to the correct answer. That is, this
prior over signal distributions guarantees that signal $i$ is more
likely to be received in state $I$ than in state $J$, but allows
signal $j$ to be more likely than signal $i$ in state $I$. The
constraint that we do place on the signal distribution allows one
to index signals and world states and make the model identifiable,
analogous to imposing an identifiability constraint to alieviate the
label switching problem when doing inference on mixture models \cite{diebolt1994estimation,jasra2005markov}. 

Respondents are modeled as Bayesians who share common knowledge \cite{samuelson2004modeling,Morris1995}
of the signal distribution and of the prior over world states. Respondents
know their received signal, but not the actual world state, as denoted
by the lack of an edge in Figure \ref{fig:basic_graphical_model-1}
between the world state node and the respondent vote node. A respondent
$r$ receiving signal $k$ has all the information necessary to compute
a posterior distribution $p(\Omega=j|T^{r}=k,S,\Psi)=p(T^{r}=k|\Omega=j,S)p(\Omega=j|\Psi)/p(T^{r}=k)$
over the world states. A respondent is assumed to wish to vote for
the world state that is most likely under their posterior distribution
over world states conditional on their received signal. For the case
of two worlds and two signals this implies that respondents receiving
signal $a$ will wish to vote for world $A$, except in the case where
the signal distribution and world prior is such that world $B$ is
more likely irrespective of the signal received. We allow for noise
in the voting data by making a respondent's vote a softmax decision
function of the posterior distribution, with the voting noise parameter
$N_{V}$ giving the temperature of this function. The voting noise
parameter is drawn from a $Gamma(3,3)$ distribution. The parameters
of this prior distribution were fixed in advance of running the model
on the datasets.

We now turn to modeling the predictions given by respondents about
the votes of other people. A respondent $r$ who received signal $k$
has the information required to compute the posterior distribution
$p(T^{s}=j|T^{r}=k)$ over the signals received by an arbitrary respondent
$s$, since $p(T^{s}=j|T^{r}=k)=\sum_{i}p(T^{s}=j|\Omega=i)p(\Omega=i|T^{r}=k)$
can be computed by marginalizing over possible worlds. Observe that
a respondent's received signal affects their predicted probability
that an arbitrary respondent receives a particular signal since how
likely an arbitrary respondent is to receive each signal depends on
the world state, but a respondent's predicted probability of the world
state depends on the signal that they received. In other words, for
respondent $r$ to compute the probability that respondent $s$ will
wish to vote for a particular world, respondent $r$can simply sum
up the probabilities of all the signals which would cause $s$ to
do this. Hence, each ideal respondent has a posterior distribution,
conditioned on their received signal, over the votes given by other
respondents. Actual respondent's predictions of the fraction of respondents
voting for $A$ are assumed to be sampled from a truncated Normal
distribution on the unit interval with a mean given by their posterior
distribution on another respondent's voting $A$ and variance $N_{M}$
(prediction noise) which is the same for all respondents and which
is sampled uniformly from $[0,.5]$. 

\subsection*{A generative possible worlds model for multiple questions }

\begin{figure}[p]
\includegraphics[scale=0.8]{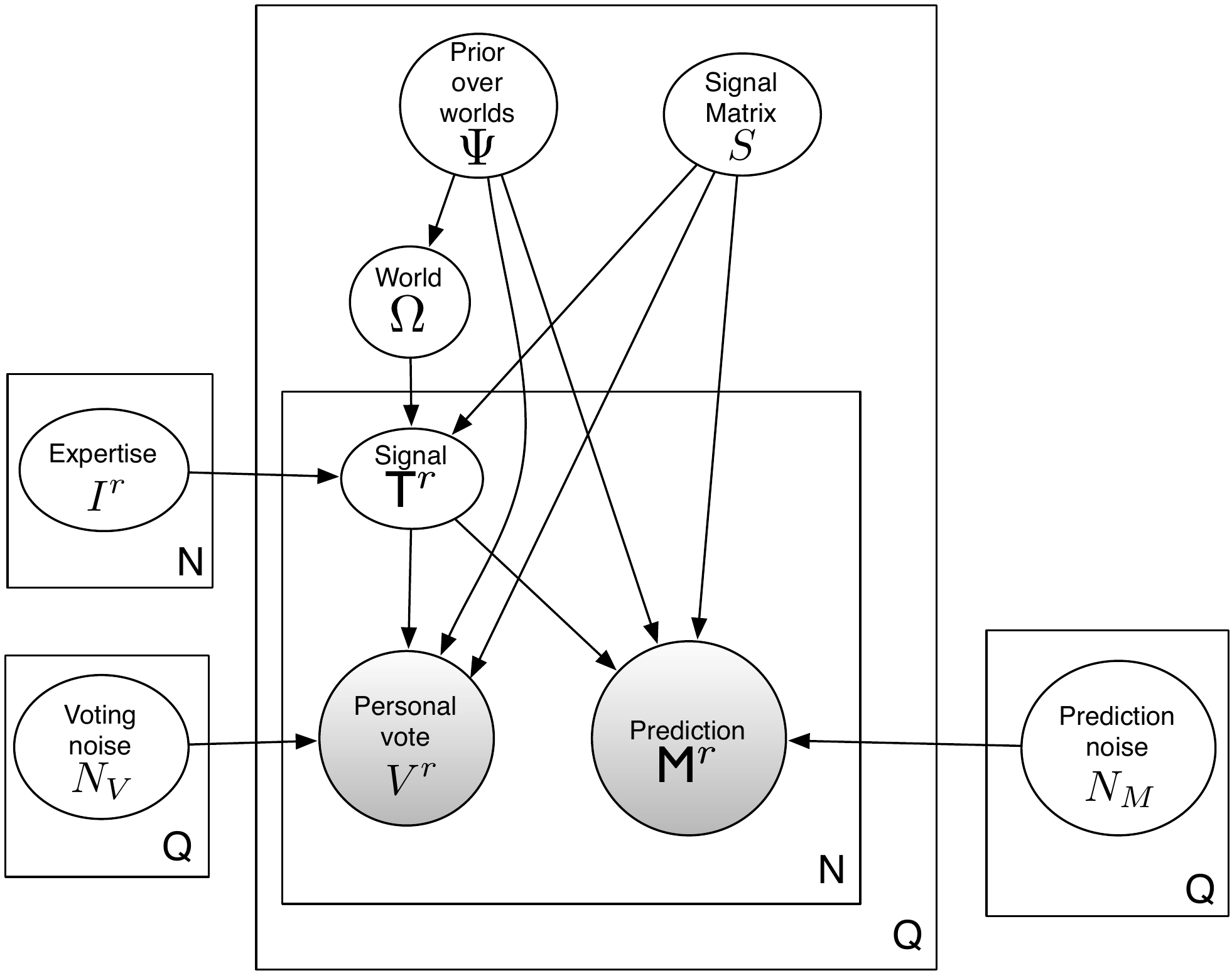}

\caption{The multiple question Possible World Model which is applied across
questions for $N$ respondents answering $Q$ questions. It is similar
to the single question model, but it includes, for each respondent,
an information expertise variable which determines how likely an individual
is to receive the correct signal compared to the baseline given by
the signal matrix.\label{fig:Complete-graphical-model}}
\end{figure}

Figure \ref{fig:Complete-graphical-model} displays a generative possible
worlds model that applies to respondents answering multiple questions.
It closely follows the single question model, but incorporates respondents'
expertise across $Q$ different questions. For each question, we sample
a world prior, world, signal distribution, and noise parameters as
with single questions, imposing no relationship across questions.
The essential difference relative to previous work \cite{lee2014using},
is that we capture differences in respondent expertise in terms of
how likely they are to receive the correct signal, rather than in
absence of error in reporting answers. We call this `information expertise'.

The information expertise parameter for respondent $r$, denoted $I^{r}$,
has support $[0,1]$ and affects how likely a respondent is to receive
the correct signal. If the probability of receiving signal $a$ in
world $A$ according to the signal distribution is $p$ (i.e. $S_{iA}=p$)
then the probability of a respondent with information expertise $I^{r}$
receiving signal $i$ is increased by $I^{r}(1-p)$ and the probability
of receiving a different signal is decreased by the same amount. That
is, the probability of receiving signal $a$ in world $A$ increases
linearly with the information expertise from the probability given
by the signal matrix to 1. For example, suppose the actual world was
$A$ and there was a 0.4 probability of receiving signal $a$ in this
world. Then, if $I^{r}=0$ the probability of respondent $r$ receiving
signal $a$ is 0.4, if $I^{r}=0.5$ then the probability of respondent
$r$ receiving signal $a$ is $0.7$, and if $I^{r}=1$ the probability
of respondent $r$ receiving signal $a$ is 1.The information expertise
parameter does not determine how accurately a respondent will answer
questions in absolute terms, but rather relative to the difficulty
of the question.For example, for an easy question where the signal
distribution gives an 80\% chance of receiving the correct signal,
even if someone has the lowest possible expertise of 0 they will still
be very likely to give the correct answer to the question. 

This model of information expertise does not allow respondents that
are less likely to receive the correct signal than the probability
given by the signal matrix. Of all the respondents, the answers of
those with expertise 0 are the most uninformative to the model.If
we additionally allowed negative expertise values between 0 and -1,
which linearly decreased the probability of receiving the signal corresponding
to the actual world, then someone with expertise -1 would always receive
the signal opposite to the actual world and so provides the same informational
content as someone with an information expertise of 1. We assume a
uniform prior distribution on $I^{r}$.\footnote{Suppose we allowed information expertise values between -1 and 1 with
a uniform prior, and our data consists only of questions where the
majority is incorrect for every question and a small subset of respondents
are correct for all the questions. Then, using the model we may infer
that the small subset of respondents have expertise values near $-1$
(and so are in a minority only because they received the wrong signal)
and that the anwer to select is the one given by the incorrect majority.
If we only allow expertise values between $0$ and $1$ then we cannot
simply infer that the minority received the incorrect signal and may
infer that the correct answer is the one endorsed by the minority,
depending on the predictions given by respondents.}

Respondents are modeled as formulating their personal votes and predictions
without taking information expertise into account. In the next section,
we develop a model which allows that respondents know their own information
expertise. 

\subsection*{Extensions of the possible worlds model\label{subsec:model_extensions}}

We test two extensions to the generative possible worlds model described
above. First, for both the single question and multiple questions
version of the model, we discuss incorporating information about the
confidence that respondents have in their answers. As for the other
elicited responses, a respondent's confidence is modeled as being
generated from the Bayesian posterior probability, conditional on
their received signal, of the answer they voted for, plus noise. That
is, if respondent $r$received signal $k$ and voted for option $j$,
their confidence is a noisy report of $p(\Omega=j|T^{r}=k)$. A respondent's
confidence is assumed to be sampled from a Normal distribution (truncated
between 0 and 1) which has a mean given by the Bayesian posterior
on the answer which they voted for, and a variance $N_{C}$ which
corresponds to the noise governing the confidence of all respondents.\footnote{For binary questions, respondents should give a confidence from 50\%
to 100\%, since if someone had less than 50\% confidence in an option
they should have voted for the alternative option. We assume a truncated
Normal distribution with support from $0$, rather than from $0.5$,
to allow for respondent voting error.} The noise $N_{C}$ is sampled from a uniform distribution on $[0,.5]$.

The second extension to the model we discuss assumes that respondents
know their own information expertise, rather than simply assuming
it is 0. In the version of the model above, a respondent's expertise
only affects their probability of receiving a particular signal, but
since everyone assumes their own expertise is 0, all respondents who
receive signal $i$ have the same posterior beliefs over worlds and
other respondents' signals. Suppose a respondent has accurate knowledge
of their own information expertise. This implies that $p(\Omega=i|T^{r}=j,I^{r}=e)=p(T^{r}=j|\Omega=i,I^{r}=e)p(\Omega=i)/p(T^{r}=j)$
where $p(T^{r}=j|\Omega=i,I^{r}=e)=S{}_{ji}+e(1-S{}_{ji})$ so that
a respondent's distribution over worlds takes into account their own
information expertise. This is in contrast to the basic version of
the model where $p(T^{r}=j|\Omega=i)=S_{ji}$ without an information
expertise term. Assuming that respondents know their own expertise
in this way has the effect that given two respondents receiving the
same signal, the one who knows that she has high information expertise
will put higher probability on the answer they endorse than the one
who believes that he has low information expertise. The respondents
will also put different probabilities on the signals received by other
respondents, since they have differing beliefs about the likelihood
of different possible worlds. A large disadvantage to assuming that
respondents know their own information expertise is the increase in
computation required for inference with this model.\footnote{We discuss inference via Metropolis Chain Monte Carlo in a later section.
If we assume that respondents know their own information expertise
then the posterior distributions over worlds and signals (conditional
on the received signal) have to be computed separately for every respondent
rather than only once, and this occurs every inference step.}As we will see in the results section, these extensions turn out not
to improve performance on the datasets from the seven studies in this
paper.

\section*{Comparison models \label{sec:Comparison-models}}

Our model can be applied to individual questions, but other generative
models for aggregation require multiple questions. To compare the
results of our model to other methods we run it on both the different
questions individually, without learning anything about individual
respondents, and also across questions to learn something about respondents
answering multiple questions. We compare our model to majority voting,
selecting the surprisingly popular answer \cite{psm}, and the linear
and logarithmic pools all of which also only require individual questions.
We also compare our model's performance to other hierarchical Bayesian
models that require multiple questions. Specifically, we compare our
model to the Lee and Danileiko cognitive hierarchy model \cite{lee2014using}
and the Bayesian Cultural Consensus model \cite{oravecz2014bayesian}.Bayesian
Cultural Consensus

Cultural Consensus Theory \cite{romney1986culture,batchelder1988test,weller2007cultural}
is a prominent set of techniques and models that are used to uncover
shared cultural knowledge, given answers from a group of people. The
theory deals with respondents answering a set of binary questions
that all relate to the same topic. Respondent's answers are used to
determine the extent to which each individual knows the culturally
correct answers (their `cultural competence') and the cultural consensus
is then determined by aggregating responses across individuals with
the answers of culturally competent people weighted more heavily.
Hence, unlike our model, cultural consensus models cannot be applied
to single questions and nor can they be applied to questions with
continuous answers. As in the Lee and Danileioko model above, the
overall assumption is thus that the consensus answer is the correct
one. The formal models of cultural consensus we consider build on
the ``General Concordet Model'' \cite{romney1986culture} and can
be thought of as factoring an agreement matrix, where each element
gives the extent to which every two individuals agree (corrected in
a particular way for guessing). 

\begin{figure}[p]
\includegraphics[scale=0.6]{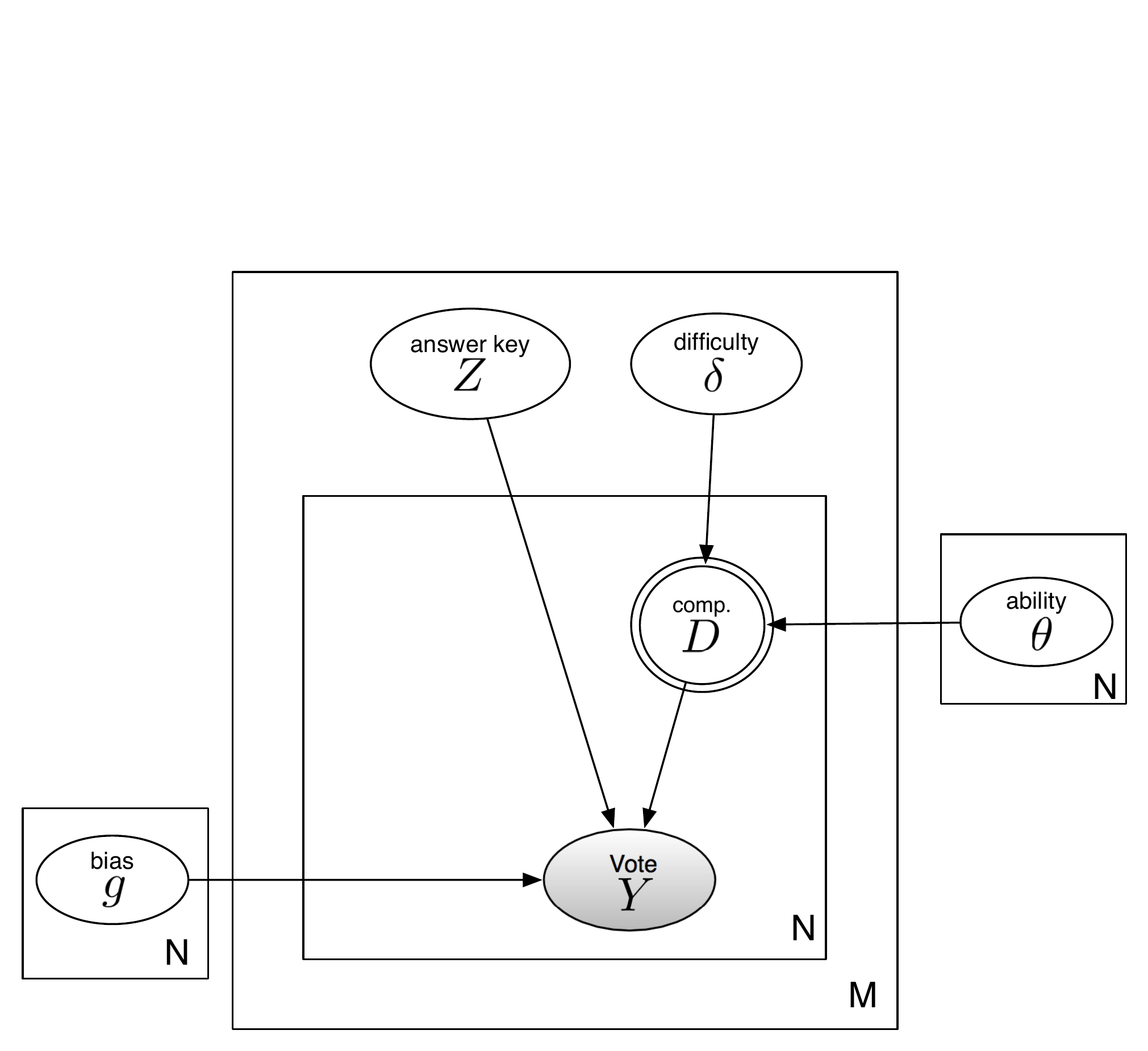}

\caption{Graphical model representation of the Bayesian Cultural Consensus
Model \cite{oravecz2014bayesian}. \label{fig:cultural-consensus-model}}
\end{figure}

In keeping with our focus on aggregation as inference, we use the
Bayesian Cultural Consensus model \cite{karabatsos2003markov,oravecz2014bayesian,oravecz2013hierarchical},
shown in Figure \ref{fig:cultural-consensus-model}, which is a generative
model based on the General Concordet Model. The model is applied to
data from $N$ respondents answering $Q$ dichotomous questions. Respondents
are indexed with $r$ and questions with $q$ and for each question
$q$, a respondent $r$votes for either true or false, denoted by
$Y_{q}^{r}\in\left\{ 0,1\right\} $. The model assumes that for each
question $q$ there is a culturally correct answer $Z_{q}\in\left\{ 0,1\right\} $.
For question $q$, a respondent $r$ knows and reports $Z_{q}$ with
probability $D_{q}^{r}$ and otherwise guesses true with probability
$g^{r}\in[0,1]$ corresponding to a respondent specific guessing-bias.
The competence $D_{q}^{r}$ of respondent $r$ at answering question
$q$ is given by the Rasch measurement model, $D_{q}^{r}=\frac{\theta^{r}(1-\delta_{q})}{\theta^{r}(1-\delta_{q})+\delta_{q}(1-\theta^{r})}$,
which is here a function of the respondent's ability $\theta^{r}\in[0,1]$
and the question difficulty $\delta_{q}\in[0,1]$. The competence
of a respondent for a question increases with the respondent's ability,
and decreases with the question's difficulty. When ability matches
difficulty the probability of the respondent knowing the answer for
the question is 0.5. Uniform priors are assumed for all parameters
of the model. The complete set of answers given by respondents can
thus be expressed as a probabilistic function of the culturally correct
answer for all questions as well as the difficulty of each question,
and the ability and guessing bias parameters of each respondent. The
model is sometimes constrained in various ways (for example, by assuming
homogeneous question difficulty, homogeneous ability across respondents,
or neutral guessing bias,) but we do not impose these constraints
and allow all parameters to vary across respondents and questions.
The posterior distribution over the culturally correct answer for
each question provides the aggregate answer, and the inferred values
of respondent ability and guessing-bias give information about a respondent's
performance.

\subsection*{A cognitive hierarchy model for combining estimates}

\begin{figure}[p]
\includegraphics{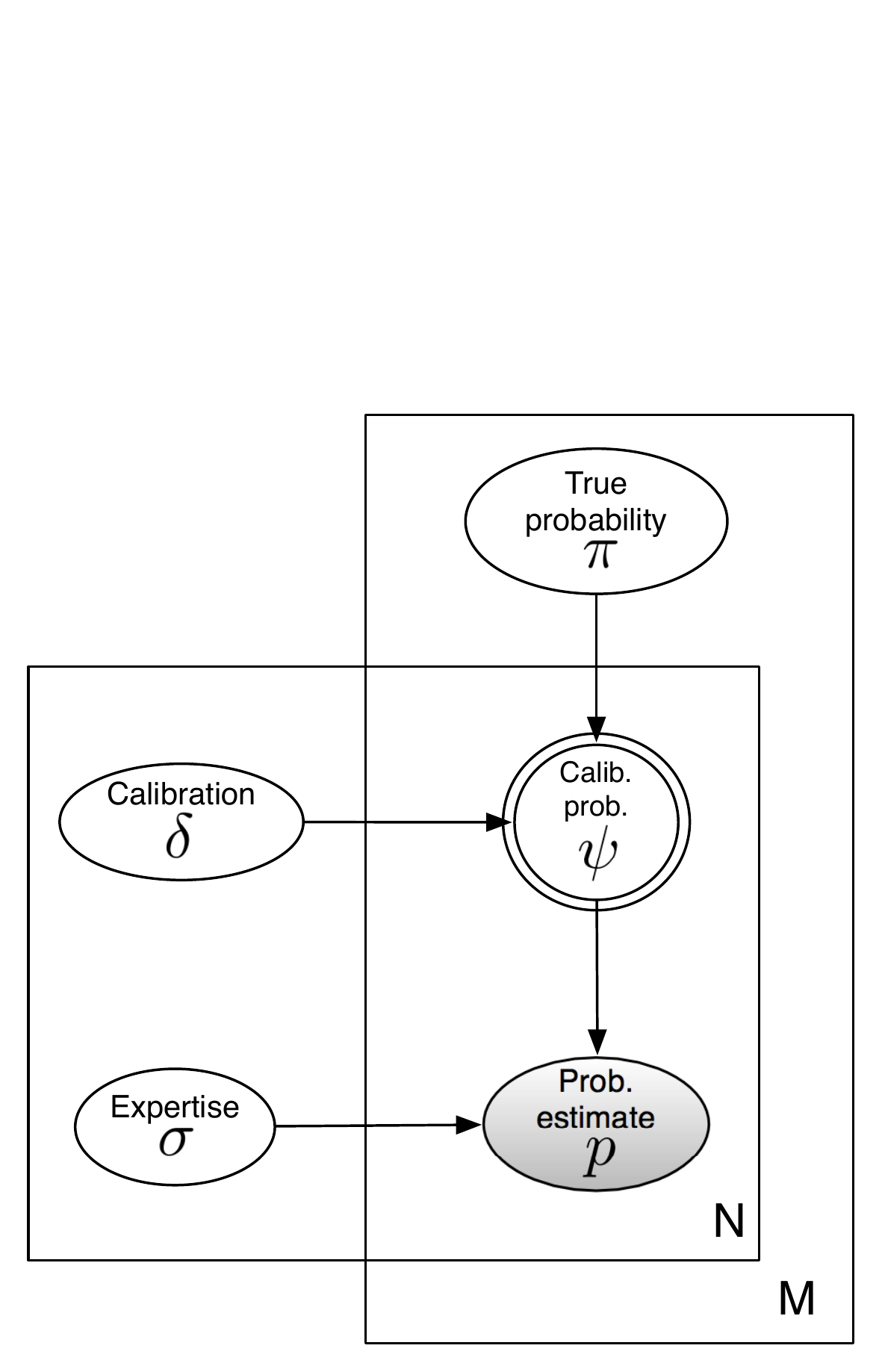}

\caption{Graphical model representation of the cognitive hierarchy model \cite{lee2014using}\label{fig:cog-hierarchy-model-1}.}
\end{figure}

We measure the performance of the model developed by Lee and Danileiko
\cite{lee2014using} which was discussed in the introduction and is
depicted in Figure \ref{fig:cog-hierarchy-model-1}. The model requires
multiple questions answered by the same set of respondents so that
individual level parameters can be learnt from the data.

The cognitive hieararchy model assumes that $N$ respondents each
answer the same set of $Q$ questions. Respondents are indexed by
$r$, questions are indexed by $q$, and the answer of respondent
$r$ to question $q$ is denoted $Y_{q}^{r}$. Respondents answers
are their subjective probabilities that the world is in a particular
state (e.g. their estimated probability that the answer to the question
is true), and so $Y_{q}^{r}\in[0,1]$. A latent true probability $\pi_{q}$
is assumed to be associated with each question and two individual-level
parameters govern how respondents report this true probability. First,
based on psychological results about how people perceive probabilities,
each respondent's perception of the true probability varies depending
on how well calibrated the respondent is. In the model, a respondent
with calibration parameter $\delta_{r}$, perceives a probability
$\psi_{q}^{r}=\delta_{r}\log(\frac{\pi_{q}}{1-\pi_{q}})$ which assumes
a linear-in-log-odds calibration function. The model further incorporates
a respondent-level parameter $\sigma_{r}$ which the authors term
`expertise' or `level of knowledge' by assuming that the reported
probability $Y_{q}^{r}$ is sampled from a Gaussian distribution centred
around the perceived probabiltiy $\psi_{q}^{r}$ with variance given
by the reciprocal of $\sigma_{r}^{2}$. That is, the larger the $\sigma^{r}$,
the more likely respondent $r$is to report a probablity closer to
their perceived probability. A uniform prior distribution on the unit
interval is assumed for both $\pi_{q}$ and $\sigma_{r}$, and the
calibration parameter for each respondent has a $Beta(5,1)$ prior. 

\section*{Evaluating the models}

\subsection*{Data}

We evaluate the models on data from running seven studies with human
participants. The seven studies were first described in \cite{psm},
and additional details about the experimental protocols for all of
these studies can be found in the supplementary information published
with the paper.\footnote{All studies were performed after approval by the MIT Committee on
the Use of Humans as Experiment Subjects, and respondents gave informed
consenst using procedures approved by the committee.} The first three studies consist of data from respondents answering
questions of the form ``Is city X the capital of state Y?'' for
every U.S. state where city X was always the most populous city in
the state. The first of these studies ($N=51$) was done in a classroom
at MIT (``MIT class states study''), the second ($N=32$) was done
in a laboratory at Princeton (``Princeton states study'') and the
third ($N=33$) was done in a laboratory at MIT (``MIT lab states
study''). Respondents indicated whether they thought the answer to
each question was true or false and predicted the fraction of respondents
answering true. For example, the first statement was ``Birmingham
is the capital of Alabama'' to which the correct response is false
because although Birmingham is Alabama's most populous city, the capital
of Alabama is Montgomery. For the third study, respondents additionally
gave their confidence of being correct on a scale from 50 percent
to 100 percent and predicted the average confidence given by others.
The prediction of the average confidence given by others is not used
in the graphical model, but we return to this kind of prediction in
the discussion section. 

Two other studies asked respondents to judge the market price of ninety
pieces of Twentieth Century art. Two groups of respondents ($N=20$
for both groups) - either art professionals, mostly gallery owners
(``Art professionals study''), or M.I.T. graduate students who had
not taken an art or art history course (``Art laypeople study'')
- worked through a paper based survey where each page contained a
color reproduction of a Twentieth Century artwork, along with information
about it dimensions and medium. Respondents indicated which of four
bins they believed a piece's price fell into: under \$1000, \$1000
to \$30 000, \$30 000 to \$1 000 000, or over \$1 000 000.\footnote{All prices in this paper refer to American dollars.}
We can collapse these answers into low and high prices with \$30 000
as the cut-off. Respondents also predicted the percentage of art professionals
and the percentage of MIT students that they believed would predict
a price over \$30 000. We only use their predictions for people in
the same group as themselves, i.e. we do not consider predictions
made by M.I.T. students about art professionals, and vice versa.\footnote{Respondents also answered questions about subjectively liking a piece
(both their own and predicting others), but we do not analyze this
data here.} The actual asking price for each piece was known to us, with 30 of
the 90 pieces having a price over \$30 000.

A sixth study had qualified dermatologists examine images of lesions
(``Lesions study'') and judge whether the pictured lesions were
benign or malignant. Respondents also predicted the distribution of
judgments on an eleven point scale which we converted to a judged
percentage, and gave their confidences on a six point Likert scale
which was likewise converted to a percentage. All the lesions had
been biopsied, and so whether each lesion was actually malignant is
known to us. For our analysis, we collapse two groups of dermatologists
one of which ($N=12$) saw a set of 40 benign lesions and 20 malignant
and the other ($N=13$) saw a set of 20 benign lesions and 40 malignant.
Thus, there were 80 images in total, half of which were benign. 

Our last study had respondents ($N=39$), recruited from Amazon's
Mechanical Turk, answer an online survey with 80 statements which
they were asked to evaluate as either true or false (``Trivia study'').
The statements were from the domains of history, science, geography,
and language. They included, for example, ``Japan has the world's
highest life expectancy'', ``The chemical symbol for Tin is Sn'',
and ``Jupiter was first discovered by Galileo Galilei''. \footnote{The correct answers are false (Monaco is higher), true, and false
(Galileo was the first to discover its four moons, but not the planet
itself. One of the earliest recorded observations is in the Indian
astronomical text, the Surya Siddhanta, from the fifth century).}For half of the questions, the correct answer was false. For each
questions, respondents indicated which answer they thought was correct,
predicted the probability that their answer was correct with a six
point Likert scale, and predicted the percentage of people answering
the survey who would answer true.

\subsubsection*{Applying the possible worlds model}

Markov Chain Monte Carlo (MCMC) inference was performed using using
the Metropolis-Hastings algorithm. The signals were marginalized out
when doing inference. We represent the world prior using the probability
of the world in state $A$, and the signal distribution matrix as
the probability of receiving signal $a$ in each state. We use truncated
Normal proposal distributions (centred on the current state with different
fixed variances for each parameter) for the world prior, noise, expertise
and signal probabilities (maintaining the constraint that the probability
of signal $a$ in state $A$ is higher than in state $B$), and we
propose the oppsite world state at each metropolis step. When doing
inference across multiple questions, the parameters for a particular
question (the signal matrix, prior over worlds, and world) are conditionally
independent of those for another question given the individual level
expertise. We thus run MCMC chains for each question in parallel with
the individual level parameters fixed, interspersed with an MCMC chain
only on the individual level parameters. For running the model on
questions separately, we use 50 000 Metropolis Hastings steps, 5000
of which are burn-in. For running the model across multiple questions,
we run 100 overall loops, the first 10 of which were burnin, where
each loop contains 2000 steps for the question parameters and 150
steps for the respondent-level parameters. We assess convergence using
the Gelman-Rubin statistic, the Geweke convergence diagnostic, and
by comparing parameter estimates across multiple chains.

\subsubsection*{Applying the Bayesian cultural consensus model}

The Bayesian Cultural Consensus Model uses only the votes of respondents.
Cultural consensus models assume that there is a unidimensional answer
key to the questions which the group is asked. A heuristic check that
the data is indeed unidimensional is to compute the ratio of the first
to the second eigenvalue of the agreement matrix with a ratio of 3:1
or higher indicating sufficient unidemnsionality \cite{oravecz2014bayesian,weller2007cultural}.
For the datasets considered in this paper the ratio of the first to
second eigenvalues were 2.76 for the MIT states-capitals class dataset,
2.62 for the Princeton states-capitals dataset, 3.32 for the MIT lab
states-capitals dataset, 2.7 for the MIT art data, 8.92 for the Newbury
art data, 2.81 for the trivia data, 10.26 for the respondents who
saw the lesions data with the split 20 malignant and 40 benign lesions,
and 6.73 for the respondents who saw the lesions data with the split
40 malignant and 20 benign lesions. Most of the datasets are higher
than the traditional standard for unidimensionality, and the model
performed well on datasets not meeting this standard. The model learns
a respondent-level guess-bias towards true. The states-capitals and
trivia questions explicitly deal with true and false answers, in the
art studies we coded true as refering to the high price option (over
\$30 000), and in the lesions study we coded true as referring to
the malignant option.\footnote{It is an open empirical question whether one would see any differences
in the responses to the question ``True or false, is this question
malignant?'' versus ``Is this lesion benign or malignant?''}

The Bayesian Cultural Consensus Toolbox \cite{oravecz2014bayesian}
specicifies the model using the JAGS model specification. This was
again altered to allow for unbalanced observational data. Gibbs sampling
was run for 1000 steps of burn-in, followed by 10 000 iterations,
using 6 independent chains and a step-size of two for thinning. As
is standard when applying MCMC to these models \cite{oravecz2014bayesian,karabatsos2003markov},
convergence was assessed using the Gelman-Rubin statistic and comparing
the parameter estimates across chains. Three of the inferred parameters
are of interest: the cultural consensus answer for each question,
the guessing bias of each respondent, and the ability of each respondent. 

\subsubsection*{Applying the cognitive hierarchy model}

The cognitive hierarchy model requires subjective probabilities from
respondents and so we apply the model only to data where this information
is available, specifically the MIT lab states study, the lesions study,
and the trivia study. We used the JAGS (Just another Gibbs sampler)
model specification provided by Lee and Danileiko with their paper,
but altered it to allow for unbalanced observational data since not
every respondent answered every question: in the MIT lab states dataset
and the trivia dataset occassionally a respondent simply missed a
question and in the lesions dataset only about half the respondents
answered some of the questions due to the experimental design. Gibbs
sampling was run for 2000 steps of burn-in, followed by another 10000
iterations, using 8 independent MCMC chains and standard measures
of autocorrelation and convergence were evaluated to ensure that the
samples approximated the posterior. We infer for each question the
latent true probability, and for each respondent their calibration
and expertise parameters.

\section*{Results}

The Bayesian cultural consensus model and cognitive hierarchy model
cannot be applied to individual questions, but our generative possible
worlds model is applied both to individual questions in a study and
also applied across questions in a study to learn respondent-level
expertise. The cognitive hierarchy model is only applied to studies
where confidences were elicited, whereas the other two models are
applied to the voting data from all seven studies. We compare the
models both with respect to their ability to infer the correct answer
to the questions and their ability to infer the expertise of respondents.
The possible worlds model allows us to infer a prior over world states
and we compare the inferred world prior to a proxy for common knowledge
about the likelihood of a city being a state capital, specifically
the frequency of mentions of city-state pairs on the Internet.

\subsubsection*{Inferring correct answers to the questions}

Of the three probabilistic models discussed, only the generative possible
worlds model can be applied to individual questions. There are other
aggregation methods, however, that can be applied to individual questions.
For all studies we compute for comparison the answer endorsed by the
majority (counting ties as putting equal probability on each answer)
and for studies where confidences were elicited we also compute a
linear opinion pool given by the mean of respondent's personal probabilities,
and a logarithmic opinion pool given by the normalized geometric means
of the probabilities that respondent's assign to each answer \cite{cooke1991experts}.
We also show the result of selecting the surprisingly popular answer
\cite{psm}, which for binary questions is simply the answer whose
actual vote frequency exceeds the average vote frequency predicted
by the sample of respondents. We discuss the results of applying the
possible worlds model both for questions separately and to all the
questions in a study together.

\begin{figure}[p]
\includegraphics[clip,scale=0.6]{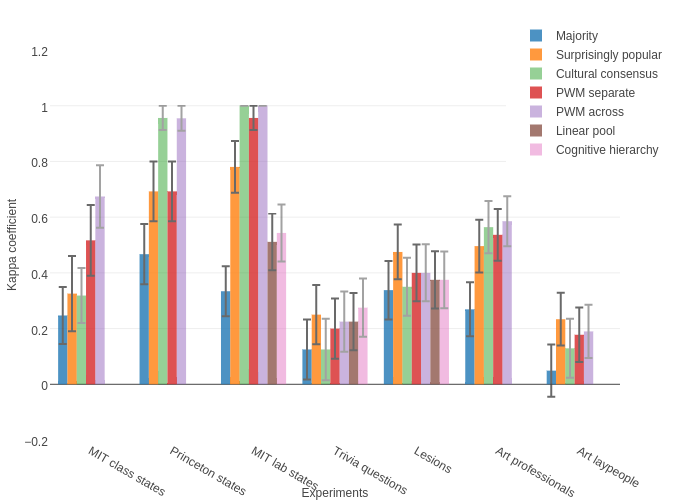}

\caption{Performance of the aggregation methods for each dataset shown with
respect to the kappa coefficient, with error bars indicating standard
errors. The lighter colored bars show methods that require data from
multiple questions. \label{fig:kappas}}
\end{figure}

\begin{figure}[p]
\includegraphics[scale=0.6]{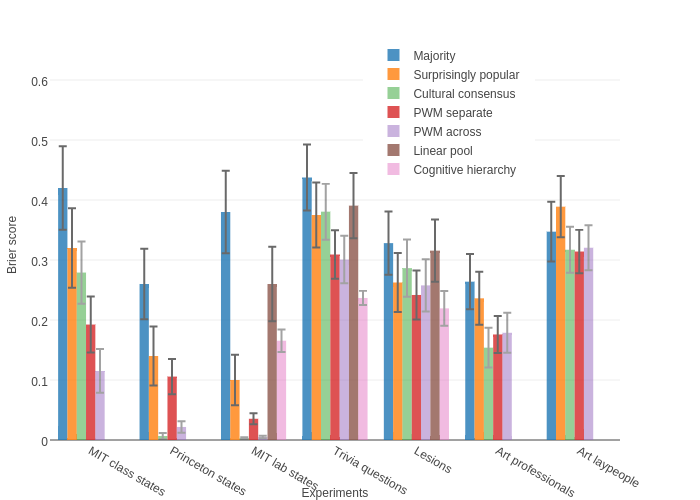}\caption{Performance of the aggregation methods for each dataset shown with
respect to the Brier score, with error bars indicating bootstrapped
standard errors. The lighter colored bars show methods that require
data from multiple questions. \label{fig:briers}}
\end{figure}

Figure \ref{fig:kappas} shows the results of each method in terms
of Cohen's kappa coefficient, where a higher coefficient indicates
a higher degree of agreement with the actual answer. Cohen's kappa
is a standard measure of categorical correlation, and we display it
rather than the percentage of questions correct since for some of
the studies the relative frequencies of the different correct answers
are unbalanced, which means that a method can have high percentage
agreement if it does not discriminate well and is instead biased towards
the more frequent answer. \footnote{Cohen's kappa is computed as $\kappa=\frac{p_{o}-p_{e}}{1-p_{e}}$
where $p_{o}$ is the relative osberved agreement between the method
and the actual answer, and $p_{e}$ is the agreement expected due
to chance, given the frequencies of the different answers output by
the method. There is not a standard technique to accomodate ties when
computing the kappa coefficient for binary questions. In the case
of ties, we construct a new set of answers with double the number
of original answers. Answers which were not ties simply appear twice
in the new set, and answers which were originally ties appear once
with one answer and once with the other answer. The kappa coefficient
is invariant to doubling the number of answers, and the standard error
is a multiple of the original number of questions and so the doubling
of the number of answers can be easily accounted for when computing
standard errors.} A disadvantage of the kappa coefficient is that it does not use the
probabilities reported by a method, but only the answer which the
method determines is more likely. Figure \ref{fig:briers} shows the
result of each method in terms of its Brier score, where a lower score
indicates that a method tends to put high probability on the actual
answer. There are a number of similar formulations of the Brier score
which we compute here as the average squared error between the probabilistic
answer given by the method and the actual answer. For methods which
output an answer rather than a probability we take the probabilities
to be zero, one, or 0.5 in the case of ties. 

We first consider the methods that act on questions individually,
which are shown with lighter bars in the two figures showing model
performance. The linear and logarithmic pools give similar answers
(the minimum kappa coefficient comparing the logarithmic and linear
pools is $0.86$), and so we do not show the logarithmic pool results
separately. Compared to the other methods that operate on questions
individually, the generative possible worlds model outperforms majority
voting, and the linear and logarithmic pool across studies if we consider
the accuracy of the answer selected by each method. This is displayed
with respect to Cohen's kappa in Figure \ref{fig:kappas}, but we
can also compare the number of errors more directly. Across all 490
items, PWM applied to questions separately improved on the errors
made by majority voting by 27\% ($p<0.001$, all p-values shown are
from a two-sided matched pairs sign test on correctness unless otherwise
indicated). Across the 210 questions on which confidences were elicited,
PWM applied to separate questions improved on the errors made by majority
vote by 30\% ($p<0.001$) and over the linear pool errors by 20\%
($p<0.02$). As can be seen, the possible worlds single question model
is able to uncover the correct answer even when run on individual
questions where the majority is incorrect. In terms of selecting one
of two binary answers, the performance of the PWM on separate questions
and the surprisingly popular answer is similar ($\kappa=0.9$ across
the 490 questions, and the answers are not significantly different
by a two-sided matched pairs sign test, $p>0.2$), which is to be
expected since they build on the same set of ideas to use people's
predictions of the distribution of answers given by the sample. However,
for all studies PWM separate question voting has a lower Brier score
than the surprisingly popular answer, since it produces graded judgments
rather than simply selecting a single answer. 

We also show the performance of models that require multiple questions
(lighter colored bars): the Bayesian cultural consensus model, the
cognitive hierarchy model, and PWM applied across questions. The PWM
across questions improves over the single question PWM ($p<0.01$),
although in terms of the kappa coefficient this improvement is small
except for two of the states-capitals studies. The cultural consensus
model also shows good performance across datasets. It is similar to
PWM across question voting in terms of the kappa coefficient, except
for the MIT states-capitals study where its performance is not as
good. This is also reflected in the difference between the correctness
of the two methods in terms of absolute numbers of questions correct
which favors the PWM applied across questions ($p=0.057)$. The cognitive
hierarchy model selects similar answers to the linear pool ($\kappa=0.92$)
resulting in similar accuracy at selecting the correct answer as measured
by Cohen's kappa, but better performance with respect to the the probability
it assigns to the correct answer (as measured by Brier score). The
cognitive hierarchy model shows similar performance to the cultural
consensus and possible worlds models on the trivia and lesions studies,
but impaired performance on the MIT lab states-capitals study.

We earlier discussed two possible extensions to the possible worlds
model: incorporating confidences and assuming that respondents are
aware of their own expertise. On the questions where confidences were
elicited, we applied the PWM with confidences incorporated both for
separate questions and across questions. Applied to questions separately,
the answers given by the PWM with and without confidences were similar
($\kappa=0.9$ on the selected answers, $r_{s}=0.87$ on the returned
probabilities). This was also the case when running the PWM across
questions with and without confidences ($\kappa=0.9$, $r_{s}=0.86)$.
Hence, incorporating confidence made little difference to the possible
world model results. We also ran the model (both with and without
confidences) assuming that people knew their own expertise. This again
made little difference to the results for either the model without
confidences ($\kappa=0.9$ for answers, $r_{s}=0.91$for probabilities)
or the model incorporating confidences ($\kappa=0.9$ for answers,
$r_{s}=0.92$ for probabilities). 

\subsubsection*{The inferred world prior and state capital mention frequency statistics }

The PWM allows one to infer the value of latent question-specific
parameters other than the world state, such as the complete signal
distribution and the prior over worlds. The accuracy of these values
is difficult to assess in general, but we analyze the inferred world
prior in the state capitals studies. Previous work in cognitive science
has demonstrated that in a variety of domains people have prior beliefs
that are well calibrated with the actual statistics of the world \cite{griffiths2006optimal}.
We use the number of Bing search results of the city-state pair asked
about in each question (specifically the search query ``City, State'',
for example ``Birmingham, Alabama'') as a proxy for how common mentions
of the city-state pair are in the world. For all three state capitals
studies, the inferred world prior on the named city being the capital
(using the PWM applied to individual questions) has a moderate correlation
with the Bing search count results under a log transform (MIT class:
$r_{S}=0.48,p<0.001$, Princeton: $r_{S}=0.49,p<0.001$, MIT lab:
$r_{S}=0.55,p<0.001$). This suggests that the model inferences about
the world prior may reflect common knowledge about how salient the
named city is in relationship to the state. 

\subsubsection*{Respondent-level parameters}

\begin{figure}[p]
\includegraphics[scale=0.7]{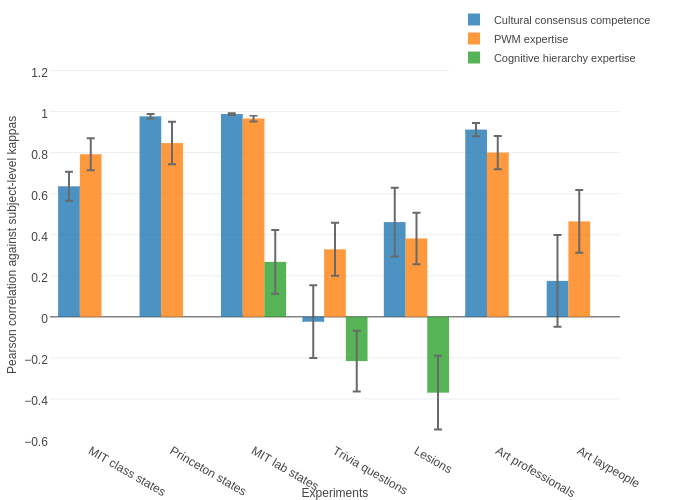}

\caption{Pearson correlations of inferred respondent level expertise parameters
from each model against the accuracy of each respondent, measured
by their kappa coefficient. Error bars show bootstrapped standard
errors. \label{fig:respondent-level-parameters}}
\end{figure}

As well as comparing results from the models to the actual answers,
we can also evaluate how well the model predicts the performance of
individual respondents. The PWM applied across questions as well as
the cultural consensus model and cognitive hierarchy model all infer
respondent-level expertise parameters. Figure \ref{fig:respondent-level-parameters}
shows how these respondent-level expertise parameters correlate with
the kappa coefficient of each respondent. For studies where confidences
were elicited, the pattern of results is the same if respondent performance
is measured using the Brier score. 

Both the PWM expertise parameter and the cultural consensus competence
parameters show high accord with individual respondent accuracy. For
the 220 total respondents in all the studies, the correlation of respondent-level
kappa accuracy with PWM expertise is $\text{r=0.79}$ (all respondent-level
correlations are significant at the $\text{p<0.005}$ level) and with
cultural consensus competence is $r=0.74$. For the 97 respondents
in the studies where confidences were elicited, the correlation of
respondent-level kappa accuracy with cognitive hierarchy expertise
is $r=0.29$, with PWM expertise is $\text{r=0.78}$ and with cultural
consensus expertise is $r=0.70$. 

For every respondent, two other pieces of information that may help
predict performance are the fraction of times that the respondent
was in the majority, and the fraction of times that the respondent
voted true. We examine the relationship between the various expertise
parameters and respondent performance if we partial out these two
additonal factors. Across all studies this partial correlation of
respondent-level kappa acurracy with PWM expertise is $r=0.85$ and
with cultural consensus competence is $r=0.74$. Across the studies
where confidence was elicited this partial correlation of respondent-level
kappa accuracy with cognitive hierarchy expertise is $r=0.34$, with
PWM expertise is $r=0.76$, and with cultural consensus competence
is $r=0.70$. 

For completeness, we also report on the other respondent-level parameters
inferred by the models. The cultural consensus guess bias parameter
correlates highly with the fraction of questions for which a respondent
says true ($r=0.95$), but not with the kappa accuracy of a respondent
($r=-0.11$, $p>.10$). The cognitive hierarchy model calibration
parameter has a correlation of $r=0.38$ with the kappa accuracy of
respondents for the studies where confidence was elicited.

\subsection*{Factors affecting model performance}

In this section, we discuss factors that affect the peformance of
the different generative models, and under what circumstances we expect
the different models to perform well. We discuss how an inconsistent
ordering across answer options affects the two models, and the role
of the predictions of others answers in the possible worlds model.

\paragraph*{A consistent coding of answers}

In these studies, the question format did not vary across items in
terms of which direction was designated as True and which as False.
If respondents have a bias toward a particular response (True or False),
this bias could be detected by some models and neutralised. However,
these questionnaires could have randomized the polarity of questions,
e.g., replacing some questions that ask whether X is true with questions
asking whether X is false. Ideally, our inference of respondent expertise
should not be affected by such inessential changes in question wording. 

To see how methods might fare with question rewording, we constructed
a new dataset where the first half of the questions in each study
are coded in reverse, and another dataset where the second half of
the questions in each study is coded in reverse. To reverse a question,
each respondent's vote is swapped (i.e. a vote for true becomes one
for false and vice versa) as is the correct answer to the question.
Additionally, a respondent's prediction of the fraction of people
voting true becomes their prediction of the fraction of people voting
false. We applied the Bayesian cultural consensus model, the cognitive
hierarchy model, and the PWM applied across questions to the half-reversed
datasets. The cognitive hierarchy model and the possible worlds model
were not affected by this transformation (the average of the kappa
coefficients for the two half-reversed datasets were within 0.05 of
the original kappas for every study). This was not the case for the
cultural consensus model, which had much lower performance on some
of the studies with the half-reversed datsets. In particular, for
the MIT class states study, the Princeton states study, and both art
studies it had a kappa of approximately $0$ for the half-reversed
datasets, in comparison to good performance on the original datasets.
This decrease in performance of the cultural consensus model is because
it relies in part on a respondent parameter which indicates the bias
towards answering true. The other models, by contrast, do not have
such a parameter and so can deal with sets of questions such that
there is not a consistent coding across questions. More generally,
the other two models can deal with sets of questions where there is
no ordering on the answer options which is the same across questions,
for example a set of questions asking which of two novel designs for
a product will be more successful. 

\paragraph*{The role of peer predictions}

The PWM uses the predictions of other people's votes to infer the
model parameters, since personal votes alone are insufficient to determine
the signal distributions in the non-actual worlds. We evaluate the
effect of these predictions on the inferences from the PWM by lesioning
the PWM to not use predictions. This lesioning of the model results
in inferences that are very similar to that given by majority voting.
Across the seven datasets, the median spearman correlation between
the inferred probability of the world being in state true by the lesioned
model and the fraction of people voting for true is $r_{S}=0.995$. 

More generally, even in situations where peer predictions are available,
these can be more or less useful depending on how accurately respondents
can give these predictions and how much variation there is across
questions - if everyone simply always predicts 50\% of people will
answer true for every question the predictions will not be useful
for improving the accuracy of the model inferences.

\section*{Discussion}

The generative PWM presented in this paper is a step towards developing
statistical methods for inferring the true world state which rely
not on the crowd's consensus answer, but rather allow for individuals
to have differential access to information. The PWM depends on assumptions
about: (1) the common knowledge that respondents share, (2) the signals
that respondents receive, and (3) the computations that respondents
make (and how they communicate them). We discuss each of these in
turn, and the possible extensions that they suggest.

\subsubsection*{Common knowledge}

One set of modeling assumptions concerns the knowledge shared by respondents.
Specifically, the PWM assumes that respondents share common knowledge
of the world prior and signal distribution. However, neither of these
assumptions are entirely correct: people neither exactly know these
quantites, and nor are beliefs about these quantities identical across
people. As discussed when comparing the inferred world prior in the
states capitals studies to Bing search results, in some cases the
world prior may reflect statistics of the environment which may be
learnt by all respondents. In other cases, expert respondents may
have a better sense of the world prior. For example, in diagnosing
whether somebody has a particular disease based on their symptoms,
knowledge of the base rate of the disease helps diagnosis but may
not be known to everyone. 

One could extend the model to weaken the common prior assumption in
various ways, although one could not simply assume that everyone had
a different belief about the prior over worlds. A model could instead
assume, for example, that respondents receiving the same signal share
a common prior over worlds, but that respondents receiving different
signals have different prior beliefs.\footnote{One would still need some kind of assumption about what respondents
believed about the prior possessed by respondents receiving other
signals.} Alternatively, one could develop models where all respondents had
noisy access to the actual world prior and signal distribution, and
formulated beliefs about other respondents knowledge of these quantities.
For example, each respondent could receive a sample from a distribution
around the actual world prior and actual signal distribution. In the
case of respondents answering many questions, one could attempt to
learn parameters that governed the accuracy of a particular respondent's
knowledge of these distributions. One could also incorporate domain
knowledge available to the aggregator into the hyperprior over world
priors. For example if one has external knowledge of the base rate
frequency of benign versus malignant lesions one could choose a hyperprior
that would make a prior matching this base rate more probable. 

\subsubsection*{Signal structure}

The PWM assumes that there are the same number of signals as world
states. It further assumes that the signals themselves have no structure:
they are simply samples from a binomial distribution (in the case
of two worlds) with signal $a$ more common in world $A$. This treats
the information or insight available to a respondent coarsely in that
it does not allow for more kinds of information available to respondents
than there are answers to a question, or for respondents with different
pieces of information to endorse the same answer. Models with more
signals than worlds could be developed with constraints on the worlds
in which different signals were more probable. For example, one could
imagine a hierarchical signal sampling process whereby each signal
was a tuple where the first element indicated the world in which the
signal was most likely and the second element gave the rank of the
probability of that signal amongst other signals with the same first
element. That is, there would be a constraint on the signal distribution
such that signal $b_{3}$, say, would be more likely in world $b$
than in any other worlds, and of the other signals more likely in
world $b$ than in any other world it was the third most probable
in world $b$. It would then be necessary to maintain these constraints
when sampling the signal distribution. More generally, models with
other kinds of assumptions about the signals that respondents receive
could be developed to more faithfully model the information available
to respondents when dealing with complex questions - for example,
respondents could receive varying numbers of signals, a mix of public
and private signals, or signals that are not simply nominal variables
but rather have richer internal structure. 

\subsubsection*{Respondent computations}

The PWM assumes that repondents can compute Bayesian posteriors over
both answers and the votes given by others, and noisely communicate
the results of their computations. The model could be extended by
modeling respondents as more plausible cognitive agents, rather than
simply as noisy Bayesians. For example, the cognitive hierarchy model
recognizes, based on much work in the psychology of decision making
\cite{zhang2012}, that respondents will be differentially calibrated
with respect to the probabilities they perceive and a similar calibration
parameter could be added to the possible worlds model for the predictions
of others answers. Modeling respondent's predictions of other people
could also incorporate what is known about this process from social
psychology. As one example, as well as showing a false-consensus effect,
people also exhibit a false-uniqueness bias such that they do not
take their own answer sufficiently into account when making their
predictions (e.g. \cite{chambers2008explaining,suls1987search}).
This could be modeled as their predictions resulting from a mixture
of their prior over signals (i.e. their knowledge of the signal distribution)
and their posterior over signals with the mixture weight given by
a respondent level false-uniqueness parameter. 

In the proposed PWM, respondents are assumed to not take some information
into account when predicting the votes given by others. We allow for
some noise when modeling how respondent's vote, but do not assume
that respondents themselves take this noise into account when predicting
the votes of others. In the model discussed here, respondents do not
attribute information expertise to other respondents, but one could
instead develop models where each respondent assumes some distribution
of information expertise across other respondents as well as modeling
the noise that they believe is present across the voting patterns
of other people.

Given the assumption that respondents compute a posterior over worlds
and over other's signals, additional statistics relating to either
of these posteriors could be elicited and modeled. For example, in
one of the states capitals studies respondents were asked to predict
the average confidence given by other respondents, which can be computed
from these two posteriors. Such additional information could potentially
help sharpen the model inferences. In practice, communicating such
questions to respondents, respondent difficulty in reasoning about
such questions, and respondent fatigue would all impose constraints
on the amount of additional data that could be elicited in this manner. 

\subsubsection*{Non-binary questions}

Lastly, we discuss extending the PWM to non-binary multiple choice
questions. Most of this extension is straightforward, since the conditional
distributions have natural non-binary counterparts. The prior over
worlds becomes a multinomial rather than binomial distribution, with
the world hyperprior drawn from a Dirichlet, rather than Beta, distribution.
The signal distribution for each world is likewise a multinomial distribution.
Respondents can still compute a Bayesian posterior distribution over
worlds and the answers of others. Respondent votes can still be modeled
with a softmax decision rule from the Bayesian posterior over multiple
possible worlds. Respondents can still compute a Bayesian posterior
distribution over the votes of others, but one needs a method of adding
noise to this posterior. One possibility is to sample from a truncated
normal distribution around each element of the posterior and then
normalize the resultant draws to sum to 1, another is to sample from
a Dirichlet distribution with a mean or mode determined from the Bayesian
posterior and an appropriate noise parameter.

\subsubsection*{Conclusion}

We have presented a generative model for inference that can be applied
to single questions and infer the correct answer even in cases where
the majority or confidence-weighted vote is incorrect. The model shows
good performance compared to models both when applied to questions
separately and when applied across multiple questions. It maintains
this performance when the answers for questions do not have a consistent
ordering. It additionally allows one to infer respondent level expertise
parameters that predict the actual accuracy of individual respondents.
While the possible worlds model that we have proposed allows multiple
extensions it is already a powerful method for aggregating the beliefs
of groups of people.

\section*{Acknowledgments}

We thank Dr. Murad Alam, Alexander Huang and Danica Mijovic-Prelec
for generous help with Study 3, and Danielle Suh with Study 4. Supported
by Sloan School of Management, Google, Inc., NSF SES-0519141, and
Intelligence Advanced Research Projects Activity (IARPA) via the Department
of Interior National Business Center contract number D11PC20058. The
U.S. Government is authorized to reproduce and distribute reprints
for Government purposes notwithstanding any copyright annotation thereon.
Disclaimer: The views and conclusions expressed herein are those of
the authors and should not be interpreted as necessarily representing
the official policies or endorsements, either expressed or implied,
of IARPA, DoI/NBC, or the U.S. Government.

\bibliographystyle{plain}
\bibliography{generative_combined}

\end{document}